# Novel Techniques to Assess Predictive Systems and Reduce Their Alarm Burden

Jonathan A. Handler, Craig F. Feied, *Senior Member, IEEE*, and Michael T. Gillam

*Abstract*— Machine prediction algorithms (e.g., binary classifiers) often are adopted on the basis of claimed performance using classic metrics such as precision and recall. However, classifier performance depends heavily upon the context (*workflow*) in which the classifier operates. Classic metrics do not reflect the realized performance of a predictor unless certain implicit assumptions are met, and these assumptions cannot be met in many common clinical scenarios. This often results in suboptimal implementations and in disappointment when expected outcomes are not achieved. One common failure mode for classic metrics arises when multiple predictions can be made for the same event, particularly when redundant true positive predictions produce little additional value. This describes many clinical alerting systems. We explain why classic metrics cannot correctly represent predictor performance in such contexts, and introduce an improved performance assessment technique using utility functions to score predictions based on their utility in a specific workflow context. The resulting utility metrics (*u-metrics*) explicitly account for the effects of temporal relationships on prediction utility. Compared to traditional measures, u-metrics more accurately reflect the real-world costs and benefits of a predictor operating in a realized context. The improvement can be significant. We also describe a formal approach to *snoozing*, a mitigation strategy in which some predictions are suppressed to improve predictor performance by reducing false positives while retaining event capture. Snoozing is especially useful for predictors that generate interruptive alarms. U-metrics correctly measure and predict the performance benefits of snoozing, whereas traditional metrics do not.

J. A. Handler is with the Clinical Intelligence Lab of OSF Healthcare, Peoria, IL 61603 USA; the Department of Emergency Medicine, Northwestern University Feinberg School of Medicine, Chicago, IL 60611; and Keylog Solutions LLC, Northbrook, IL 60062 USA (email: jhandler@gmail.com).

C. F. Feied is with the Department of Emergency Medicine, Georgetown University School of Medicine, Washington, DC 20007 USA; Asatte Consulting Associates LLC, Honolulu, HI 96816 USA; and WhisperSom Corporation, Annapolis, MD 21409 USA (e-mail: craig.feied@asatte.net).

M. T. Gillam is with HealthLab, Washington, DC 20003 USA (email: mike@healthlab.com).

## I. INTRODUCTION

Medicine is increasingly adopting artificial intelligence and machine learning (ML) techniques to predict future events that may be susceptible to clinical intervention. An approach commonly used in medicine is *Boolean classification*, in which the probability of a future event is predicted and the prediction is classified as *positive* or *negative* using a selected cutoff. A positive prediction means the event is predicted to occur during a subsequent time interval (the "*prediction window*"). To assess the performance of the classifier, each prediction is compared to the observed outcome and scored as true or false. The counts of true and false positive and negative predictions are used to compute statistical measures of predictor performance including recall, precision, and other count-based metrics (*c-metrics*) such as those shown in Table III(a) [1].

C-metrics have well-established clinical and practical utility in a wide variety of contexts. However, they derive solely from the counts of positive and negative predictions that are true or false, giving equal weight to each prediction. For this reason, they cannot adequately measure the realized performance of Boolean predictors *in praxi* unless three implicit assumptions (the *required classical assumptions*) are met. First, utility is *dichotomized*: all benefit comes from true predictions and all adversity comes from false predictions. Second, utility is *uniformly distributed*: every prediction has the same relative value, thus simple counts of predictions are a valid proxy for the utility. Third, the utility for each prediction is *symmetric*: the realized benefit or harm from each prediction has precisely the same magnitude as the missed benefit or avoided harm that would have resulted had the opposite (*complementary*, *potential alternative*) prediction been made.

A workflow scenario in which prediction utility is dichotomized, uniformly distributed, and symmetric is a *conforming* workflow. A workflow scenario in which prediction utility is incompletely dichotomized, nonuniformly distributed, or asymmetric is a *nonconforming* workflow. When the required classical assumptions hold true, prediction counts are an accurate measure of relative utility. Otherwise, they are not. The required classical assumptions fail in many common clinical situations. **When the required classical assumptions do not hold true, traditional c-metrics do not correctly reflect classifier performance.**

Failure of the required classical assumptions does not arise from the classifier itself. Rather, it arises from attributes of the population and from the way in which a classifier is deployed and utilized within a workflow. As a simplified example, consider a classifier that predicts clinical decompensation and is implemented in two different workflow scenarios.

In Scenario One, the classifier runs within an order entry system and automatically admits patients either to an ICU bed or to a regular bed, depending upon whether a prediction of decompensation is positive or negative. False positives incur unnecessary costs and false negatives may result in catastrophic clinical outcomes. False negatives thus cause loss of potential benefit and *also* represent a new source of additional harm.

In Scenario Two, each admitting physician decides whether their patient should go to an ICU or a regular bed. Subsequently, the admitting physician is notified if the classifier predicts decompensation for a patient assigned to a regular bed, offering an opportunity to change the admission to an ICU bed. In this scenario false negatives may result in loss of a potential benefit, but they *do not* represent a new source of additional harm.

The classifier makes the same predictions in the two workflows, and the correctness of each prediction is the same. Traditional classifier performance measures derive solely from counts of true and false positive and negative predictions, thus c-metric performance scores will be the same in the two scenarios despite the fact that the impact on cost and clinical outcomes is competely different.

*Why count-based metrics fail*

In c-metrics, simple counts of predictions serve as a proxy for utility, thus the relative utility $u$ of each individual prediction is always assumed to be $u = 1$. This explains why traditional performance measures often do not match the *realized performance* experienced by users. C-metrics reflect the theoretical performance of a classifier operating *in a conforming workflow*, and cannot account for the nonuniform, nondichotomized or asymmetric prediction utility that often characterizes real-world workflows. We shall see that the application of utility coefficients to c-metrics can correct for certain types of nonuniform utility but not for others, particularly those that violate dichotomy or symmetry.

NONUNIFORM DISTRIBUTIONS OF UTILITY

C-metric performance measures will not match realized predictor performance when the real distribution of utility is nonuniform. For example, c-metrics such as *sensitivity* or *recall* penalize a predictor equally for each false negative across all workflow scenarios, even when false negatives cause serious harm in one scenario and minimal or no harm in another. Similarly, c-metrics reward a predictor equally for all true predictions, resulting in misleadingly high performance scores when true negatives have less value than true positives (e.g., when only positive predictions trigger notifications or when class imbalance is high).

C-metrics also yield misleadingly high performance scores when the utility of true positive predictions varies over time, as it often does in clinical workflows. For example, an *impending seizure* alert that leads clinicians to administer a short-acting anticonvulsant has low utility if the drug is given too far in advance, or if the prediction occurs (or the notification arrives) too late to respond before the seizure. Utility distribution may depend on small details of the workflow (e.g., the definition of "too late" will be different depending on whether the drug is kept on the hospital unit or must be ordered from a central pharmacy). Prediction utility may vary if users cannot always act upon the prediction (e.g., predictions that are made after closing time, or on the weekend). Although this example illustrates a nonuniform distribution of utility over time, utility distributions may be nonuniform with respect to any variable.

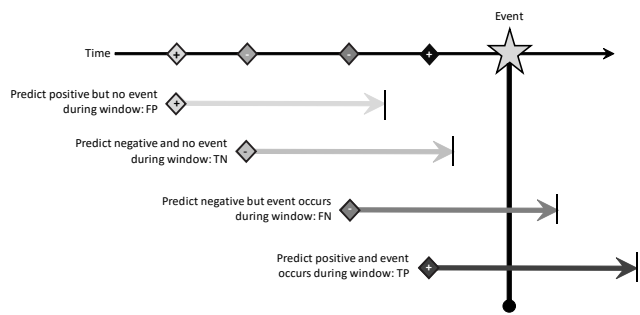

Fig. 1. Scoring with overlapping prediction windows. Multiple true and false predictions may be made for the same event, and prediction windows may overlap more than one event. When a prediction's utility depends upon its timing in relation to an event, each prediction window is truncated at the first event encountered. Other solutions are possible, but truncation has the advantage that training and performance analysis match the way the classifier operates in practice.

One common source of utility nonuniformity arises when multiple predictions occur for the same event, producing overlapping prediction windows as shown in Fig. 1. Longer prediction windows and shorter prediction intervals may be advantageous for several reasons. One advantage is that longer windows result in higher effective prevalence for each prediction window. Another is that more frequent predictions can detect events earlier (if predictions for the next 24 hours are made only once daily at midnight, an event occurring at 12:01 AM can be correctly predicted at best one minute in advance). Another is that in some machine learning algorithms, more frequent predictions increase the likelihood that transient (*sentinel*) signals will naively contribute to predictive patterns. For example, before it was known that transient atrial fibrillation independently predicts mortality after cardiac bypass surgery [4], a machine learning system could "learn" the pattern by making predictions frequently enough to capture the fleeting occurrences.

The fact that a given classifier makes predictions having overlapping prediction windows may not be immediately obvious. For example, many cardiac monitors contain a classifier that issues an alarm each time a couplet of paired ventricular ectopic beats is detected. Couplet alarms predict the onset of more serious arrhythmias (e.g., with cardiac arrest) within the next few hours [3]. Many such alarms may be issued in the hours preceding a serious arrhythmia. The clinical utility of the classifier is entirely associated with the serious arrhythmia outcome, and multiple prediction windows overlap with respect to this outcome.

When multiple true positive and false negative predictions may be made for the same realized event, the utility of a prediction often depends upon relative timing. For example, when positive predictions generate alarms, more frequent predictions may lead to more frequent alarms. The first true positive prediction creates value, but what about subsequent true positive predictions for the same event? Alarm fatigue is a well-known problem: two false alarms are worse than one, and even true alarms can be disruptive if redundant alarms are triggered too frequently [5], [6].

When redundant true positive predictions cause alarms that generate negative utility rather than positive utility, the distribution of utility is both nonuniform (predictions do not all have the same utility) and non-dichotomized (some true alarms produce adverse utility). However, c-metrics always assign positive utility of 1 to *every* true prediction.

NONDICHOTOMIZED PREDICTION UTILITY

C-metrics performance measures reward a predictor for every true prediction and penalize it for every false prediction, thus will not match realized performance if any true predictions create adverse utility or any false predictions create beneficial utility. For example, c-metrics do not fit observed performance if multiple predictions are made for the same event and redundant notifications (alarms) result from true positive predictions yet produce negative utility.

ASYMMETRIC PREDICTION UTILITY

Traditional c-metrics also assume *symmetry* of prediction utility (i.e., the realized utility $u$ of each prediction is assumed to be the same magnitude as the complementary utility $u_c$ that would have resulted had the opposite prediction been made). Since the two are assumed to be identical, complementary utility is not tabulated explicitly; for measures that depend on complementary utility, c-metrics uses *realized utility*. For this reason, c-metrics will not reflect realized predictor performance when prediction utility is asymmetric. Asymmetric prediction utility is frequently encountered in real-world workflows. For example, in screening scenarios (such as Scenario Two, above) a false negative prediction that creates no additional harm has negligible realized utility ($u$) but high complementary utility ($u_c$), representing the missed benefit of the prediction.

The absence of elements representing complementary utility is a serious shortcoming of c-metrics. Without such elements, c-metrics lacks a robust formula for calculating the fraction of available benefit captured (*recall*). C-metrics uses the true positive prediction rate as a proxy for recall, considering the two to be the same. However, this substitution is valid only when the utility of a true positive and its potential alternative false negative are the same (i.e., when prediction utility is symmetric). This constraint is rarely met in clinical scenarios. Failures of c-metrics due to nonuniform utility distributions that violate the symmetry requirement cannot be fixed through the simple addition of utility coefficients.

*A novel solution*

We present a novel solution addressing the root causes of the mismatch between c-metric performance measures and the realized performance of Boolean predictors, together with a novel technique that can improve realized predictor performance in certain common scenarios.

UTILITY-BASED METRICS

Utility based metrics (*u-metrics*) parallel traditional count-based metrics but derive from a new *utility-based confusion matrix*. U-metrics do not use counts as a proxy for utility, do not assume that each prediction has the same value, and do not assume that all true predictions are beneficial or all false predictions are harmful. They do not assume that the utility of a prediction is the same as the potential utility of the complementary prediction that *could* have been made. They operate equally in conforming and nonconforming workflows, and in conforming workflows they yield the same results as c-metrics. They correctly measure performance even when utility is incompletely dichotomized, nonuniformly distributed, or asymmetric. They are resilient under class imbalance. They include robust formulae for metrics that include potentially available benefit (e.g., recall). U-metrics permit accurate assessment of classifier performance in situations where c-metrics fail because the *required classical assumptions* do not hold.

Utility-based scoring differs from that of traditional count-based c-metrics in three fundamental ways that directly address the shortfalls of c-metrics with respect to *incomplete dichotomization*, *nonuniformity*, and *asymmetry*.

1) *Dichotomization*: In c-metrics each prediction is scored *True* or *False*; all benefit comes from true predictions while all adversity comes from false predictions. In u-metrics, predictions are scored based on the *beneficial* or *adverse utility* generated, where utility is a *scalar* quantity having magnitude without sign. Utility scores are entered into a *utility-based confusion matrix* that is organized by type of utility without respect to prediction truth. This allows u-metrics to handle situations in which utility is not dichotomized.
2) *Uniformity*: In c-metrics each prediction has an implied relative utility of 1 (i.e., simple counts of predictions are the measure of relative utility). In u-metrics the relative utility of each prediction is estimated based on the context in which it is made. This allows u-metrics to handle situations in which utility is nonuniformly distributed.
3) *Symmetry*: In c-metrics, *realized utility* is represented in the confusion matrix as counts of true and false predictions, but *complementary utility* is not tabulated. In u-metrics, both realized and complementary utility are explicitly tabulated in the confusion matrix. This allows u-metrics to handle situations in which prediction utility is *asymmetric*.

SNOOZING

We will also introduce a formal approach to the temporary suppression of notifications generated by positive predictions from a Boolean classifier (*snoozing*). Snoozing can reduce false alarms and redundant true alarms, improving the precision-recall curve across a range of predictor cutoffs. In many cases, snoozing can significantly improve precision with no loss of recall. Utility-based metrics properly account for snoozing whereas count-based metrics cannot.

ALARM-CENTRIC SCENARIOS

Both snoozing and u-metrics will be shown to be especially useful in *alarm-centric scenarios*, a class of common nonconforming workflows having specific characteristics:

- Multiple predictions can be made for the same event.
- Positive predictions trigger interruptive notifications.
- Initial notifications are more useful than subsequent redundant notifications for the same event.
- False positive predictions produce adverse utility.
- True negative predictions produce very little utility.

Although scenarios exist in which true negative predictions generate significant utility, in many contexts the utility of true negative predictions is low enough to be disregarded. For example, negative predictions may not generate alerts; users may not be aware of negative predictions at all, and may assume rare events will not occur unless notified otherwise. The meaning of a negative prediction may be ambiguous, as when the user is unaware of classifier internals and does not know whether a negative prediction means the event is strongly predicted not to occur, or weakly predicted to occur (e.g., just below the cutoff for a positive prediction).

## II. Related Work

Not uncommonly, the calculated performance metrics for a predictive classifier fail to match the realized benefits obtained when the predictor is deployed in the real world [2]–[5]. To identify related work, a literature search was performed using PubMed, Cochrane Library, ArXiv, Semantic scholar, Google Scholar, and Google. Combinations of the following search terms were used: Machine (learning, prediction, classifier), Boolean/Binary (classification, classifier, prediction, predictor), Performance, Precision, Recall, Receiver Operating Characteristic, AUC, PPV, Confusion matrix, Utility (adjusted, weighted, coefficient, based), Alert/Alarm fatigue, Snooze, Snoozing. Each document found was reviewed for relevance.

The search strategy yielded a number of domain-specific examples of mismatch between calculated performance metrics and realized performance of Boolean predictors, along with corrections for various types of utility nonuniformity.

Examples were found in many domains of endeavor. In 1884, Peirce (economics) described the "utility of the method" using an average profit across all true positive predictions and an average loss across all false positive predictions [6]. Drummond (engineering technology) described the use of fixed cost functions to account for asymmetric misclassification costs [7]. Kruchten (data visualization) added fixed utility coefficients to a traditional confusion matrix in the calculation of expected benefit [8]. Borghetti and Gini (metareasoning in computer science) noted that prediction errors may incur nonuniform costs [9]. Rass (cybersecurity) noted that utility functions may not be uniform with respect to true and false predictions and that false predictions may not always produce adverse utility [10]. Gordon et al. (human-computer interface design) noted that machine learning classifiers for human-facing tasks often score highly on traditional performance metrics but are received poorly in practice, and developed a denoising deconvolution transformation to correct for certain types of utility nonuniformity common in that domain [2]. Dal Seno et al. (brain-computer interfaces) noted that information transfer rate (ITR) accurately measures the performance of a classifier in isolation but poorly describes the overall performance of real systems, and proposed as an alternative the ratio between the average benefit of a choice (assigned using a utility function for each prediction) and the average time required to achieve it [5]. Lobo et al. (ecology and geographic biosampling) identified 5 problems that make the area under the receiver operating characteristic curve (AUC) poorly suited as a performance measure, including the fact that AUC values omission and commission errors equally, whereas in many applications the two do not have the same importance [3].

In the domain of medicine, Jung et al. identified the mismatch between standard performance metrics and realized performance as an important reason why machine learning models have failed to produce desired clinical and economic outcomes in practice. They described the addition of fixed utility coefficients to each cell of a traditional count-based confusion matrix, noted that utility may not necessarily be fixed, and applied post-hoc adjustments to account for specific types of nonuniform utility distributions [11]. Tomasev (medicine) proposed the use of recall calculated per clinical event with precision calculated per six-hour evaluation period, presumably in an attempt to avoid overvaluing redundant predictions made for the same clinical event [12].

We found no mention of temporary suppression of notifications generated by positive predictions from a Boolean classifier (*snoozing*) in any domain. McCoy and Das (medicine) use the word snooze to describe reducing prediction frequency to one per 6-hour period [13]. Horvitz (computer science) describes a method to delay notifications until a moment when the utility of the alert can be maximized for the user [14]. Although we found strategies for mitigation of sensor alarm fatigue, each was based on the manipulation of alarm thresholds (e.g., adaptive threshold adjustments) and resulted in the traditional tradeoffs between precision and recall [15]. We found no suggestion of snoozing to reduce alarm burden while improving predictor precision at a given recall.

### *Summary of related work*

Many others have recognized the mismatch between traditional performance metrics and the realized performance of Boolean predictors. Several authors describe corrections for utility nonuniformity that may be useful in specific situations. However, to the best of our knowledge the fundamental reasons for the discrepancy between traditional performance metrics and realized performance have not previously been explicated, nor has a general utility-based solution previously been set forth. We found no prior mention of snoozing to improve Boolean classifier performance, nor of the implications of *asymmetric prediction utility*, nor of *complementary utility* and its use in the calculation of various types of recall.

## III. Methods A: Utility-Based Metrics

### *Scoring and calculating utility-based metrics*

Table I contrasts the counts of the traditional count-based confusion matrix (*c-matrix*) with the utility scores of the utility-based confusion matrix (*u-matrix*). C-matrix columns indicate outcome (true or false) whereas u-matrix columns indicate the type of utility generated (beneficial or adverse).

Functions to estimate the relative utility of each prediction in a workflow typically are developed by a deployment team in collaboration with stakeholders. Predictions may create either beneficial or adverse utility. Any range of scalar values may be chosen to represent utility, but the range from 0 to 1 is preferred because it is computationally convenient.

Utility is assigned according to scoring functions $f$ in which the utility ($u$) typically is a function of the workflow context in which the prediction is made ($w$) as well as the state of the system ($s$), the outcome ($o$), and the prediction itself ($p$) so that $u = f(w,s,o,p)$. Since nonuniform utility may vary with respect to multiple variables, arbitrarily complex scoring functions are possible. For any series of variables $v_1 .. v_N$ the scoring function may be expressed as $u = f(v_1..v_N)$. At times, the development of acceptable utility functions may require clinical investigation, expert opinion, and stakeholder negotiations.

In practice, most utility functions are simple and derive from clinician perceptions of value (e.g., "assign beneficial utility of 1.0 to the first true positive prediction and utility of 0 to redundant positive predictions for the same event," or "assign each true positive prediction a beneficial utility of 1.0 when the prediction can be acted upon and 0 when it cannot.").

## COMPLEMENTARY UTILITY

The u-matrix elements of Table I(b) represent utility that was realized from the predictions that were made (*realized utility*). These elements correspond directly to the elements of a c-matrix, and are sufficient for the calculation of u-metrics that derive solely from realized utility, such as u-precision. However, other metrics such as u-recall must also take into account the total amount of benefit or harm that *could* have been captured had all predictions been made in such a way as to maximize that benefit or harm. Such metrics require additional elements to represent *complementary utility* ($u_c$), the missed benefit (complementary benefit $B_C$) or avoided harm (complementary adversity $A_C$) that would have resulted if the opposite prediction had been made. If the realized utility of a prediction is beneficial, the complementary utility will be adverse, and vice-versa. If either is zero, the other determines whether the zero is treated as beneficial or adverse. Utility scoring functions are written only for *realized utility*; the *complementary utility* of a prediction is the realized utility that would have resulted from the opposite prediction.

## UTILITY-BASED CONFUSION MATRIX

Table II shows the complete utility-based confusion matrix (u-matrix) containing both *realized utility* elements and the additional *complementary utility* elements required for calculation of metrics that derive from potentially available benefit or potentially avoidable harm, such as the *captured utility* metrics of Table III(c).

The u-matrix is fundamentally different from a c-matrix because it characterizes predictions by utility independent of correctness, it allows for arbitrary amounts of benefit or adversity (including zero) from any prediction, and it explicitly accounts for asymmetric valuation of realized and complementary utility.

## REALIZED VS CAPTURED POTENTIAL METRICS

The realized and complementary elements of Table II form the basis of the quantitative *realized utility* and *captured potential utility* u-metrics shown in Table III(b) and (c). Realized utility metrics derive solely from the utility of predictions as made, whereas captured potential

TABLE I
COUNT-BASED VS UTILITY-BASED CONFUSION MATRICES

|  | True Predictions | False Predictions |
|---|---|---|
| Predict Positive | TP | FP |
| Predict Negative | TN | FN |

Legend:
**TP**: count of true positive predictions
**TN**: count of true negative predictions
**FP**: count of false positive predictions
**FN**: count of false negative predictions

|  | Beneficial Utility | Adverse Utility |
|---|---|---|
| Predict Positive | BP | AP |
| Predict Negative | BN | AN |

Legend:
**BP**: summed utility of Beneficial Positives
**AP**: summed utility of Adverse Positives
**BN**: summed utility of Beneficial Negatives
**AN**: summed utility of Adverse Negatives

(a) Count-based confusion matrix    (b) Realized utility confusion matrix

(a) The traditional count-based confusion matrix has been trivially rearranged to show true predictions in one column and false predictions in the other. This arrangement helps in visualizing the distinction between c-metrics and u-metrics. In c-metrics, all benefit comes from true predictions and all adversity comes from false predictions.

(b) The columns of a utility-based confusion matrix indicate whether a prediction produces *beneficial* or *adverse* utility, regardless of prediction truth.

TABLE II
COMPLETE UTILITY-BASED CONFUSION MATRIX

|  | Complementary Adverse Utility | Realized Beneficial Utility | Realized Adverse Utility | Complementary Beneficial Utility |
|---|---|---|---|---|
| Predict Positive | $A_C(BP)$ | BP | AP | $B_C(AP)$ |
| Predict Negative | $A_C(BN)$ | BN | AN | $B_C(AN)$ |

| BP | ***Beneficial Positive Utility.*** The realized utility of Beneficial Positive Predictions |
|---|---|
| $A_C(BP)$ | ***Complementary Adverse Utility of Beneficial Positives.*** The potential alternative utility of Beneficial Positive predictions had the opposite prediction been made. |
| AP | ***Adverse Positive Utility.*** The realized utility of Adverse Positive Predictions. |
| $B_C(AP)$ | ***Complementary Beneficial Utility of Adverse Positives.*** The potential alternative utility of Adverse Positive predictions had the opposite prediction been made. |
| BN | ***Beneficial Negative Utility.*** The realized utility of Beneficial Negative predictions. |
| $A_C(BN)$ | ***Complementary Adverse Utility of Beneficial Negatives.*** The potential alternative utility of Beneficial Negative predictions had the opposite prediction been made. |
| AN | ***Adverse Negative Utility.*** The realized utility of Adverse Negative Predictions. |
| $B_C(AN)$ | ***Complementary Beneficial Utility of Adverse Negatives.*** The potential alternative utility of Adverse Negative predictions had the opposite prediction been made. |

Utility-based confusion matrix with both *realized* and *complementary* elements. Complementary elements represent missed benefit or avoided harm.

TABLE III
COUNT-BASED AND UTILITY-BASED PERFORMANCE METRICS

| Traditional count-based metrics | | "Realized Utility" utility-based metrics | | "Captured Potential" utility-based metrics | |
|---|---|---|---|---|---|
| True Positive Rate (Sensitivity, Recall) | $\frac{TP}{TP+FN}$ | **u-Sensitivity** u-Beneficial Positive Rate | $\frac{BP}{BP+AN}$ | **u-Recall** u-Beneficial Positive Capture Rate | $\frac{BP}{BP+B_C(AN)}$ |
| True Negative Rate (Specificity) | $\frac{TN}{TN+FP}$ | **u-Specificity** u-Beneficial Negative Rate | $\frac{BN}{BN+AP}$ | u-Beneficial Negative Capture Rate | $\frac{BN}{BN+B_C(AP)}$ |
| False Positive Rate | $\frac{FP}{FP+TN}$ | u-Adverse Positive Rate | $\frac{AP}{AP+BN}$ | u-Adverse Positive Capture Rate | $\frac{AP}{AP+A_C(BN)}$ |
| False Negative Rate | $\frac{FN}{FN+TP}$ | u-Adverse Negative Rate | $\frac{AN}{AN+BP}$ | u-Adverse Negative Capture Rate | $\frac{AN}{AN+A_C(BP)}$ |
| Positive Predictive Value (Precision) | $\frac{TP}{TP+FP}$ | **u-Precision** u-Positive Predictive Value | $\frac{BP}{BP+AP}$ | u-Benefit Capture Rate for Positive Predictions | $\frac{BP}{BP+B_C(AP)}$ |
| Negative Predictive Value (Negative Precision) | $\frac{TN}{TN+FN}$ | u-Negative Precision u-Negative Predictive Value | $\frac{BN}{BN+AN}$ | u-Benefit Capture Rate for Negative Predictions | $\frac{BN}{BN+B_C(AN)}$ |

(a) Traditional c-metrics    (b) Realized utility metrics    (c) Captured potential utility metrics

(a) C-metrics reflect realized performance only in conforming workflows. (b) Realized utility u-metrics are used when measuring the effect of predictions as made. (c) Captured potential u-metrics are used when measuring the fraction of available utility captured. They may be thought of as a series of recall metrics reflecting the fraction captured of one type of utility relative to the total (maximum) potential utility of that type.

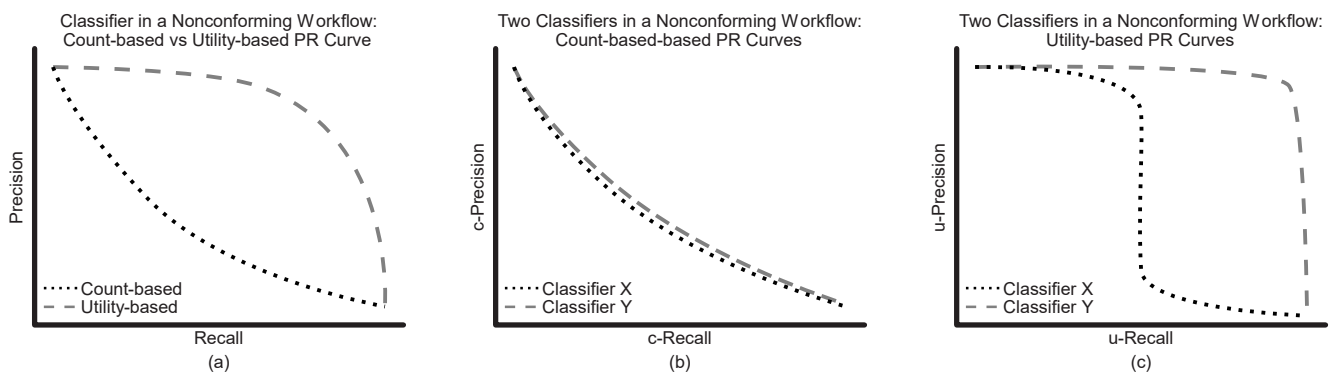

Fig. 2. (a) the count-based c-PR curve is depressed and has low AUC-PRC due to penalization for missed redundant predictions that would not have added value. The utility-based u-PR curve correctly reflects the actual clinical utility with higher AUC-uPRC. (b) Count-based c-PR curves may be virtually identical for two classifiers (X,Y) with dramatically different practical performance. (c) Utility-based u-PR curves correctly distinguish between the same two classifiers.

utility metrics include the utility potentially available if all predictions had been made in such a way as to maximize that type of utility.

*Sensitivity and recall*

As illustrated in Table III, the fraction of events correctly predicted (traditionally *sensitivity*) and the fraction of total benefit captured (traditionally *recall*) are distinct concepts. In conforming workflows where c-metrics are valid, a simple count of false negatives represents both the count of missed events and the missed benefit that could have been captured if the false negatives had been positive predictions, thus the calculation $TP/(TP+FN)$ yields both sensitivity and recall. Since c-metrics uses the same formula for both, sensitivity and recall often are conflated and treated as a single measure.

However, the two measures have the same value only when the utility of every prediction is *symmetric*: if the utility of each prediction and that of its opposite alternative prediction are not precisely equal under all scenarios, then *sensitivity and recall will not have the same value*. In most real-world workflows prediction utility is not symmetric; thus, the concept of beneficial positive rate (sensitivity) should not be conflated with the concept of captured potential utility (recall). Inability to separate these two measures is an important shortcoming of c-metrics that cannot be mitigated by adjustments to correct for utility nonuniformity.

*Choosing between metrics*

If prediction utility is symmetric, each *realized utility* metric will produce the same numeric result as its corresponding *captured potential utility* metric. When prediction utility is not symmetric, the choice of which Table III metric to use depends on what is being measured. When measuring the effect of predictions that were made, a *realized utility* metric should be selected. When measuring the success of a classifier in capturing benefit or avoiding harm, a *captured potential utility* metric should be used. When considering the probability that any single positive or negative prediction will be correct, the traditional positive predictive value (c-PPV) or negative predicted value (c-NPV) should be selected. When considering the *expected benefit* of a single positive or negative prediction, the u-PPV or u-NPV should be selected. When considering the success of the predictor in capturing benefit through positive or negative predictions, the u-beneficial rate for positive or negative predictions should be selected.

An additional measure of interest is the ratio AP/BP (*Adversity Ratio*) showing how much adverse effect must be tolerated per unit of beneficial effect from positive predictions. The adversity ratio is analogous to the c-metric False/True ratio, which tells how many false positives must be tolerated for every true positive [12]. The adversity ratio correctly accounts for the fact that predictions made in different contexts and at different times may create different amounts of benefit or harm.

As with any statistical measure, thoughtfulness is required when determining whether or not a specific utility-based metric is applicable to a given situation. If scoring rules or functions cause any element in the u-matrix to be always zero, derived measures that comprise that element may reduce to values of zero, one, or undefined, making them less useful.

*Tradeoff curves and area under the curve*

Boolean classifiers often are described and compared on the basis of two curves: the receiver operating characteristic (*ROC*) curve, which in c-metrics shows the tradeoff between true positive rate (sensitivity) and false positive rate, and the precision-recall (*PR*) curve, which in c-metrics shows the tradeoff between the true positive rate (recall) and the positive predictive value (precision) [16], [17].

ROC curves are strongly affected by true negative predictions, making them poor descriptors in alarm-based scenarios or whenever events to be predicted are rare [3], [18].

Precision-recall curves properly account for class imbalance, but count-based PR curves may fail in nonconforming workflows because they falsely assume that all predictions have the same relative value. For example, Fig. 2(a) illustrates a finding that is all too common in alarm-based scenarios: the count-based c-PR curve (dotted) is artificially depressed by redundant false negative predictions that yield a low calculated c-recall *even if all events are successfully detected* (i.e., the calculated c-recall for the predictor is low, whereas the user-experienced recall of events is 100%). In contrast, the utility-based u-PR curve (dashed) is based on u-matrix elements that explicitly account for asymmetry and nonuniform utility across predictions, resulting in u-PR curves that correctly match realized classifier performance.

Fig. 2(b) and (c) illustrate an extreme example in which two classifiers each make ten predictions per event in an alarm-centric scenario. At a specified cutoff, classifier Y makes five correct (TP) and five incorrect (FN) predictions for each event, thus *every event is detected*. Classifier X makes ten correct (TP) predictions per event for half of the events and ten incorrect (FN) predictions per event for the other half, thus *half the events are missed*. At each cutoff, the two classifiers produce nearly identical counts of TP, TN, FP, and FN, thus c-PR curves and AUC-cPRC will be virtually identical for the two classifiers even though at some cutoffs one classifier correctly identified every event whereas the other missed half of the events. In contrast, utility-based u-PR curves and AUC-uPRC correctly reflect the observed performance difference between the two classifiers. Fig. 2(b) and (c) illustrate this phenomenon.

## IV. METHODS B: UTILITY-BASED EXAMPLES

Two clinical deployment scenarios (scenario A, scenario B) illustrate the use of typical scoring functions. The worked example for scenario B appears in Table IV, Table V, and Table VI.

**EXAMPLE SCENARIO A**

A limited number of isolation rooms are allocated by a classifier that predicts the likelihood of active tuberculosis. During implementation, stakeholders decided that in this scenario every prediction is equally important: true negatives and true positives are valued equally, and it is equally important to avoid false positives and false negatives. The scoring rules are summarized in Fig. 3(a).

*Scenario A Rules:*
1. Under all circumstances:
   Every true prediction has a beneficial utility of 1.0.
   Every false prediction has an adverse utility of 1.0.

*Scenario A Discussion:*

The stakeholder-determined utility functions satisfy the requirements underlying traditional count-based metrics, thus the utility-based confusion matrix and derived u-metrics are equivalent to the traditional c-metrics, and are calculated in the usual manner.

**EXAMPLE SCENARIO B**

A classifier makes predictions every ten seconds. A loud alarm is sounded each time the classifier predicts impending cardiac arrest within the next five minutes. Utility functions were determined by an institutional QA committee based on a pilot study. Since the cost and disruption of mobilizing a response team is high, the harm of a false positive was judged to be the same magnitude as the benefit of a true positive. All false positives were considered to be equally harmful, since each one could trigger a new response. Because of their disruptive potential, redundant true positives were considered 20% as harmful as false positives. Negative predictions do not cause an alert, and the committee judged that with

| (a) Scenario A scoring rules | (b) Scenario B scoring rules |
|---|---|
| **In all cases**<br>• True Positives<br>  $BP = A_C(BP) = 1.0$<br>• True Negatives<br>  $BN = A_C(BN) = 1.0$<br>• False Positives<br>  $AP = B_C(AP) = 1.0$<br>• False Negatives<br>  $AN = B_C(AN) = 1.0$ | **If at least one true positive for event:**<br>• First True Positive = 1.0 BP; $A_C$(BP) 0.0<br>• Extra True Positives = 0.2 AP; $B_C$(AP) 0.0<br>• False Negatives = 0.0 BN; $A_C$(BN) 0.2<br>**If no true positives for event:**<br>• First False Negative = 0.0 AN; $B_C$(AN) 1.0<br>• Extra False Negatives = 0.0 BN; $A_C$(BN) 0.2<br>**In all cases**<br>• True Negatives = 0.0 BN; $A_C$(BN) 1.0<br>• False Positives = 1.0 AP; $B_C$(AP) 0.0 |

Fig. 3. (a) Scenario A scoring results in utility-based performance metrics that are identical to count-based metrics. (b) Scenario B utility scoring reflects a nonconforming workflow with realized and complementary utility.

TABLE IV
SCENARIO B: TABULATED RESULTS OF PREDICTIONS

| Prediction | Actual Outcome | Event # | Classic Assessment | Realized Utility | Complementary Utility |
|---|---|---|---|---|---|
| **Negative** | Positive | 1 | False Negative | 0.0 BN | 0.2 $A_C$(BN) |
| **Positive** | Positive | 1 | True Positive | 1.0 BP | 0.0 $A_C$(BP) |
| **Positive** | Positive | 1 | True Positive | 0.2 AP | 0.0 $B_C$(AP) |
| **Negative** | Positive | 1 | False Negative | 0.0 BN | 0.2 $A_C$(BN) |
| **Positive** | Negative | -- | False Positive | 1.0 AP | 0.0 $B_C$(AP) |
| **Negative** | Negative | -- | True Negative | 0.0 BN | 1.0 $A_C$(BN) |
| **Negative** | Positive | 2 | False Negative | 0.0 AN | 1.0 $B_C$(AN) |
| **Negative** | Positive | 2 | False Negative | 0.0 BN | 0.2 $A_C$(BN) |

In this scenario having 2 events, only one is detected. Of three alarms triggered, only one creates value (the first true positive for the first event).

TABLE V
SCENARIO B: RESULTING COUNT-BASED VS UTILITY-BASED CONFUSION MATRICES

| Traditional count-based Confusion Matrix | | | Utility-based Confusion Matrix | | | | |
|---|---|---|---|---|---|---|---|
| | True Prediction | False Prediction | | Complementary Adverse Utility | Beneficial Utility | Adverse Utility | Complementary Beneficial Utility |
| Predict Positive | 2 (TP) | 1 (FP) | Predict Positive | 0.0 $A_C$(BP) | 1.0 BP | 1.2 AP | 0.0 $B_C$(AP) |
| Predict Negative | 1 (TN) | 4 (FN) | Predict Negative | 1.6 $A_C$(BN) | 0.0 BN | 0.0 AN | 1.0 $B_C$(AN) |

TABLE VI
SCENARIO B: CALCULATION OF COUNT-BASED VS UTILITY-BASED METRICS

| | Traditional count-based metrics | | | "Realized Utility" utility-based metrics | | | "Captured Potential" utility-based metrics | | |
|---|---|---|---|---|---|---|---|---|---|
| True Positive Rate (Sensitivity, Recall) | $\frac{TP}{TP+FN}$ | $\frac{2}{2+4} = 0.33$ | **u-Sensitivity** u-Beneficial Positive Rate | $\frac{BP}{BP+AN}$ | $\frac{1.0}{1.0+0.0} = 1.00$ | **u-Recall** u-Beneficial Positive Capture Rate | $\frac{BP}{BP+B_C(AN)}$ | $\frac{1.0}{1.0+1.0} = 0.50$ |
| True Negative Rate (Specificity) | $\frac{TN}{TN+FP}$ | $\frac{1}{1+1} = 0.50$ | **u-Specificity** u-Beneficial Negative Rate | $\frac{BN}{BN+AP}$ | $\frac{0}{0+1.2} = 0$ | u-Beneficial Negative Capture Rate | $\frac{BN}{BN+B_C(AP)}$ | $\frac{0.0}{0.0+0.0} =$ Undef |
| False Positive Rate | $\frac{FP}{FP+TN}$ | $(\frac{1}{1+1}) = 0.50$ | u-Adverse Positive Rate | $\frac{AP}{AP+BN}$ | $\frac{1.2}{1.2+0} = 1.00$ | u-Adverse Positive Capture Rate | $\frac{AP}{AP+A_C(BN)}$ | $\frac{1.2}{1.2+1.6} = 0.43$ |
| False Negative Rate | $\frac{FN}{FN+TP}$ | $(\frac{4}{4+2}) = 0.67$ | u-Adverse Negative Rate | $\frac{AN}{AN+BP}$ | $\frac{0.0}{0.0+1.0} = 0.00$ | u-Adverse Negative Capture Rate | $\frac{AN}{AN+A_C(BP)}$ | $\frac{0.0}{0.0+0.0} =$ Undef |
| Positive Predictive Value (Precision) | $\frac{TP}{TP+FP}$ | $\frac{2}{2+1} = 0.67$ | **u-Precision** u-Positive Predictive Value | $\frac{BP}{BP+AP}$ | $\frac{1.0}{1.0+1.2} = 0.45$ | u-Benefit Capture Rate for Positive Predictions | $\frac{BP}{BP+B_C(AP)}$ | $\frac{1.0}{1.0+0.0} = 1.00$ |
| Negative Predictive Value (Negative Precision) | $\frac{TN}{TN+FN}$ | $\frac{1}{1+4} = 0.20$ | u-Negative Precision u-Negative Predictive Value | $\frac{BN}{BN+AN}$ | $\frac{0}{0+0.0} =$ Undef | u-Benefit Capture Rate for Negative Predictions | $\frac{BN}{BN+B_C(AN)}$ | $\frac{0}{0+1.0} = 0$ |

normal clinical vigilance, negative predictions create no benefit or harm. The resulting scoring rules are summarized in Fig. 3(b).

***Scenario B Rules:***

1. <u>For predictions with windows that intersect an event where at least one true positive prediction exists:</u>

   - Since an event occurs, there can be no false positives or true negatives.
   - The first true positive prediction has realized utility of 1.0 BP and complementary utility of 0 $A_C(BP)$ (because the opposite prediction would have been a false negative).
   - Redundant true positives for the event are assigned a utility of 0.2 AP, reflecting a small adverse effect for each distracting excess alert. The complementary utility is assigned a value of 0 $B_C(AP)$.
   - False negatives for the event are assigned a realized utility of zero (extra reminders were unwanted, thus neither harm nor benefit is realized from missed reminders). If the alternative (positive) prediction had been made, it would have been a redundant true positive with adverse utility of 0.2 (as above). The complementary utility is accordingly assigned a value of 0.2 $A_C(BN)$, which means the realized utility of zero is treated as beneficial.

2. <u>For predictions with windows that intersect an event where no true positive predictions exist:</u>

   - Since an event occurs, there can be no false positives or true negatives.
   - False negatives for the event have realized utility of 0 AN because the absence of an alert does not create new harm.
   - The first false negative prediction has potential alternative utility of 1.0 $B_C(AN)$ because if positive it would have been the first true positive.
   - Each subsequent false negative has potential alternative utility of 0.2 $A_C(BN)$ because if positive it would have been an unwanted redundant alarm.

3. <u>For predictions with windows that do not intersect an event:</u>

   - Since an event does not occur, there are no true positives or false negatives.
   - Each true negative prediction has realized utility of 0 BN and potential alternative utility of 1.0 $A_C(BN)$.
   - Each false positive has utility of 1.0 AP and potential alternative utility of 0 $B_C(AP)$.

***Scenario B Discussion:***

The sample dataset for Scenario B contains 8 predictions and 2 events. The worked example is shown in Table IV, Table V, and Table VI. This example illustrates a common situation in which traditional c-metrics misrepresent the true performance of the classifier through overscoring or underscoring. Overscoring occurs when a count-based metric improperly benefits from elements that do not increase the practical benefit of the classifier. Underscoring occurs when a count-based metric is improperly penalized for elements that do not reduce the practical benefit of the classifier.

The c-metric *c-precision* is falsely elevated (67%) because it gives full value to both the first and the second true positive alert for event #1, when the second alert is merely an annoyance. The u-metric *u-precision* (45%) more accurately reflects the clinical reality: of three positive predictions one was incorrect (did not capture an event), one was correct and captured an event, and one was a redundant prediction that added no benefit and was interruptively annoying.

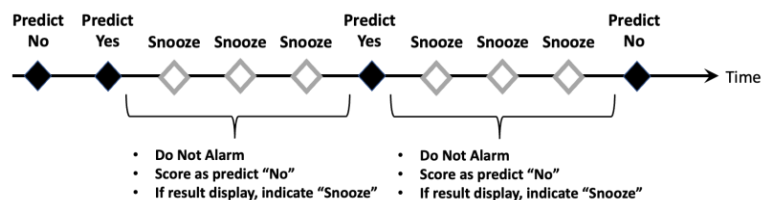

Fig. 4. Snoozing acts to suppress any alarms that would be triggered by predictions during the snooze period. This improves u-metrics that incorporate adverse positive utility, such as u-precision, u-specificity, and the adversity ratio.

Similarly, the c-metric *c-recall* is falsely low (33%) because it receives a full penalty for every false negative, even though false negatives cause no harm so long as at least one true positive captures the event. The captured potential *u-recall* metric correctly shows that 50% of the potential benefit (one of the two events) was captured by the classifier.

## V. METHODS C: IMPROVING CLASSIFIER PERFORMANCE

When false positive predictions trigger notifications (alarms), they may cause harm by increasing cost and effort, worsening alarm fatigue, and triggering clinical interventions [19]. Up to 80–99% of clinical alarms are false alarms [20]. In many cases, two special techniques can be applied to improve realized classifier performance and reduce false alarms. The first technique is the use of longer prediction windows, and the second is the use of snoozing to achieve notification pruning. The performance improvements obtained are correctly characterized by u-metrics but not by c-metrics.

### *Longer Prediction Windows*

In alarm-centric scenarios, increasing prediction window duration can improve classifier performance by increasing the effective prevalence of the event in the prediction windows. U-metrics correctly reflect the real-world utility resulting from such a change, but c-metrics may not. For example, if overlap increases, each event will be associated with more predictions and c-metrics will blindly penalize the classifier for redundant false negatives while incorrectly rewarding the classifier for redundant true positives.

### *Notification Pruning by Snoozing*

Clinical classifiers often issue repeated positive predictions in response to conditions that persist over time. In *alarm-centric scenarios* a temporary suppression of notifications (*snoozing*) can improve realized performance by eliminating unwanted redundant alarms arising from such prediction clusters.

In alarm-centric scenarios, snoozed notifications are experienced as if they had been negative predictions, thus are assigned the same utility as negative predictions (typically zero in an alarm-centric scenario). As long as snoozing does not suppress every true positive prediction for an event, this results in significantly improved uPR curves with better u-precision, u-specificity, and adversity ratio for any given level of u-recall.

Snoozing may be initiated and terminated based on any defined criteria. For example, *time-based snoozing* is a strategy whereby snoozing is automatically activated by a positive prediction, after which alerts are suppressed for a defined time period, resuming at the end of that period, as illustrated in Fig. 4. Other strategies may also be useful. For example, snoozing may begin after a positive prediction and continue until either a positive prediction with greater certainty is made or the prediction window ends. This approach may be useful, for example, in cases where the patient has entered a new steady state and additional alarms at the prior level of certainty would have low utility. Multiple snooze strategies may be combined, and snoozing may be combined with other techniques such as adaptive thresholds [15] to further reduce alarm burden.

U-metrics properly account for the performance benefits of snoozing, but c-metrics cannot do so because any attempt to ignore or reclassify snoozed predictions results in unacceptable side effects leading to meaningless performance measures.

### SCENARIO C WITH TIME-BASED SNOOZING

Snoozing is illustrated by scenario C, based on Scenario B with the addition of time-based snoozing. The worked example is shown in Table VII, Table VIII, and Table IX. Predictions are made at the start of every 10-minute interval. A snooze time of 40 minutes is used, and snoozing is automatically started when a positive prediction is made.

TABLE VII
SCENARIO C WORKED EXAMPLE WITH AND WITHOUT SNOOZING

| Time | Prediction | Actual | Event ID | Classic Count-based Scoring | Count-based with Snoozed Predictions Treated as Negative | | Utility-Based Scoring With No Snooze | | | Utility-Based Scoring With 40-minute Snooze | | |
|---|---|---|---|---|---|---|---|---|---|---|---|---|
| 0 minutes | Negative | Negative | -- | True Negative | TN | | 0.0 BN | 1.0 $A_C$(BN) | | 0.0 BN | 1.0 $A_C$(BN) | |
| 10 minutes | Negative | Negative | -- | True Negative | TN | | 0.0 BN | 1.0 $A_C$(BN) | | 0.0 BN | 1.0 $A_C$(BN) | |
| 20 minutes | Positive | Negative | -- | False Positive | FP | | 1.0 AP | 0.0 $B_C$(AP) | | 1.0 AP | 0.0 $B_C$(AP) | |
| 30 minutes | Positive | Negative | -- | False Positive | TN | Snoozed | 1.0 AP | 0.0 $B_C$(AP) | | 0.0 BN | 1.0 $A_C$(BN) | Snoozed |
| 40 minutes | Positive | Negative | -- | False Positive | TN | Snoozed | 1.0 AP | 0.0 $B_C$(AP) | | 0.0 BN | 1.0 $A_C$(BN) | Snoozed |
| 50 minutes | Positive | Negative | -- | False Positive | TN | Snoozed | 1.0 AP | 0.0 $B_C$(AP) | | 0.0 BN | 1.0 $A_C$(BN) | Snoozed |
| 60 minutes | Negative | Negative | -- | True Negative | TN | | 0.0 BN | 1.0 $A_C$(BN) | | 0.0 BN | 1.0 $A_C$(BN) | |
| 70 minutes | Negative | Positive | 1 | False Negative | FN | | 0.0 BN | 0.2 $A_C$(BN) | | 0.0 BN | 0.2 $A_C$(BN) | |
| 80 minutes | Positive | Positive | 1 | True Positive | TP | | 1.0 BP | 0.0 $A_C$(BP) | | 1.0 BP | 0.0 $A_C$(BP) | |
| 90 minutes | Negative | Positive | 1 | False Negative | FN | Snoozed | 0.0 BN | 0.2 $A_C$(BN) | | 0.0 BN | 0.2 $A_C$(BN) | Snoozed |
| 100 minutes | Positive | Positive | 1 | True Positive | FN | Snoozed | 0.2 AP | 0.0 $B_C$(AP) | | 0.0 BN | 0.2 $A_C$(BN) | Snoozed |
| 110 minutes | Positive | Positive | 2 | True Positive | FN | Snoozed | 1.0 BP | 0.0 $A_C$(BP) | | 0.0 BN | 0.2 $A_C$(BN) | Snoozed |
| 120 minutes | Positive | Positive | 2 | True Positive | TP | | 0.2 AP | 0.0 $B_C$(AP) | | 1.0 BP | 0.0 $A_C$(BP) | |

In this scenario having 2 events, there are 2 wanted alarms (one for each event) and 6 unwanted alarms: 4 false positives and 2 redundant true positives. Of these, classic scoring recognizes only the 4 false positives as unwanted. Utility-based scoring assigns an Adverse Positive score to each of the 6 unwanted alarms. The addition of snoozing reduces the unwanted alarm count from 6 to 1.

TABLE VIII
SCENARIO C: RESULTING CONFUSION MATRICES WITH AND WITHOUT SNOOZING

| | Traditional Count-based Confusion Matrix | | Count-based With Snoozed Predictions Treated as Negative | | | Utility-based Confusion Matrix Without Snoozing | | | | Utility-based Confusion Matrix With Snoozing | | | |
|---|---|---|---|---|---|---|---|---|---|---|---|---|---|
| | True | False | True | False | | Complementary Adverse | Realized Beneficial | Realized Adverse | Complementary Beneficial | Complementary Adverse | Realized Beneficial | Realized Adverse | Complementary Beneficial |
| Predict Positive | 4 (TP) | 4 (FP) | 2 (TP) | 1 (FP) | Predict Positive | 0.0 $A_C$(BP) | 2.0 BP | 4.4 AP | 0.0 $B_C$(AP) | 0.0 $A_C$(BP) | 2.0 BP | 1.0 AP | 0.0 $B_C$(AP) |
| Predict Negative | 3 (TN) | 2 (FN) | 6 (TN) | 4 (FN) | Predict Negative | 3.4 $A_C$(BN) | 0 BN | 0.0 AN | 0.0 $B_C$(AN) | 6.8 $A_C$(BN) | 0 BN | 0.0 AN | 0.0 $B_C$(AN) |

**If at least one true positive for event:**
- 1.0 BP for first true positive ($A_C$(BP) 0.0)
- 0.2 AP for additional true positives ($B_C$(AP) 0.0)
- 0.0 BN for false negatives ($A_C$(BN) 0.2)

**If no true positives for event:**
- 0.0 AN for first false negative ($B_C$(AN) 1.0)
- 0.0 BN for additional false negatives ($A_C$(BN) 0.2)

**In all cases**
- 0 BN for all true negatives ($A_C$(BN) 1.0)
- 1.0 AP for all false positives ($B_C$(AP) 0.0)

**During snoozing:**
- All snoozed predictions are scored as if negative

Fig. 5. Scoring rules for Scenario C.

TABLE IX
SCENARIO C: RECALL AND PRECISION WITH AND WITHOUT SNOOZING

| Count-based metrics | | | | Utility-based metrics | | | |
|---|---|---|---|---|---|---|---|
| C-metric | | Without Snoozing | Snoozed Treated as Negative | U-metric | | Without Snoozing | With Snoozing |
| Recall | $\frac{TP}{TP+FN}$ | $\frac{4}{4+2}=0.67$ | $\frac{2}{2+4}=0.33$ | Recall | $\frac{BP}{BP+B_C(AN)}$ | $\frac{2.0}{2.0+0.0}=1.0$ | $\frac{2.0}{2.0+0}=1.0$ |
| Precision | $\frac{TP}{TP+FP}$ | $\frac{4}{4+4}=0.50$ | $\frac{2}{2+1}=0.67$ | Precision | $\frac{BP}{BP+AP}$ | $\frac{2.0}{2.0+4.4}=0.31$ | $\frac{2.0}{2.0+1.0}=0.67$ |

Utility metrics correctly describe classifier performance and correctly show the effects of snoozing. In this scenario, snoozing improves u-precision from 31% to 67% while u-recall is maintained at 100%. Count-based metrics cannot benefit from snoozing.

*Scenario C Rules:*

Utility scores are assigned as in scenario B, with the addition of the simple snoozing rule as shown in Fig. 5. Since positive predictions during the snooze period are experienced as if they were negative predictions (i.e., they do not generate alerts) they are scored as if they were negative predictions. If the prediction window overlaps an event they are scored as false negatives, otherwise they are scored as true negatives. Their complementary utility is the utility that would have been captured if the prediction had been experienced as positive (i.e., an alarm had been generated).

*Scenario C Discussion:*

The worked example corresponds to a common clinical scenario in which c-metrics do not match realized performance, snoozing improves realized performance, and c-metrics cannot measure the performance improvement obtained by snoozing.

*Without snoozing*

The dataset for scenario C contains thirteen predictions and two events. With no snoozing, the predictor detects both events, thus the *clinically meaningful* recall as experienced by the user is 100%. U-recall reflects this correctly, but the corresponding c-recall undervalues the classifier at 67% because it fully penalizes *clinically irrelevant* false negative predictions for events that were detected by at least one true positive prediction. C-metrics also overvalue classifier precision at 50%, because full credit is assigned to redundant true positives that produce no clinical value and are experienced as interruptive nuisances. In contrast, the u-precision metric at 31% correctly accounts for the fact that two redundant true positives reduced clinical utility rather than increasing it.

*With snoozing*

The addition of snoozing improves classifier performance by reducing adverse positive (false positive and redundant true positive) predictions while still capturing every event. U-metrics correctly reflect this improvement: u-recall remains unchanged (at 100%) while u-precision improves significantly (from 31% to 67%). C-metrics cannot properly account for the performance improvements obtained through snoozing because there is no acceptable way to handle snoozed predictions. They cannot simply be ignored because c-recall would not be penalized for completely missing an event, resulting in meaningless performance metrics. They cannot simply be reclassified as negative, either: although in this example c-precision would improve, snoozed redundant true positive predictions would be treated as false negatives, worsening c-recall (from 67% to 33%). This makes no sense, because in reality 100% of events were captured.

Improper use of c-metrics to assess the potential benefits of snoozing leads to the false perception that any improvement in precision comes at the expense of recall, making it appear as if similar "tradeoffs" could be achieved simply by adjusting the classifier cutoff. In fact, the real-world performance benefits of snoozing can be substantial, cannot be replicated by adjustment of cutoffs, and are correctly described by u-metrics but not by c-metrics. We believe this explains a lack of prior interest in snoozing for Boolean predictors.

*Classifier Optimization Through Ranging*

In an alarm-centric scenario, classifier performance often can be improved through careful choices of cutoff and snooze duration. Setting the cutoff too high or the snooze duration too long can cause a predictor to miss events completely. A practical method for optimization is as follows: for each snooze duration of interest, the classifier cutoff is varied and a *ranging* table is generated with the observed performance metrics of interest at each cutoff. From all the rows of all the ranging tables, a combination of snooze duration and classifier cutoff are chosen that best achieve the desired parameters of performance for the use case (e.g., emphasizing recall while maintaining a specified minimum precision, or vice versa).

TABLE X

| No Snoozing | C-metric | U-metric | Reality Check |
|---|---|---|---|
| AUC-PR | .24 | .18 | |
| Recall | .47 | .50 | 50% of events were captured |
| Precision | .29 | .20 | 18% of alarms were for a new event |
| | | | 23 unwanted alarms occurred |

(a) Cutoff 0.275; No Snooze

| Snooze 24 h | C-metric | U-metric | Reality Check |
|---|---|---|---|
| AUC-PR | .24 | .32 | |
| Recall | .29 | .50 | 50% of events were captured |
| Precision | .31 | .31 | 31% of alarms were for a new event |
| | | | 11 unwanted alarms occurred |

(b) Cutoff 0.25; Snooze 24 hrs

Snoozing improves realized performance while reducing unwanted alarms. U-metrics correctly match user experience, but c-recall is falsely reduced.

## VI. REAL-WORLD EXPERIMENT AND RESULTS

To illustrate the use of u-metrics and snoozing with a clinical dataset, the Catboost algorithm [22] was used to develop a predictive model for hypertensive events occurring within a 24 hour window following each blood pressure measurement. The research plan was reviewed by the institutional review board of the University of Illinois College of Medicine at Peoria.

The model was developed using a publicly available clinical dataset comprising a time series of blood pressures and related variables (e.g., time of day, heart rate, and association with meals) [21]. The dataset is class-imbalanced for hypertensive events (SBP > 140) with about 14% of rows in the positive class.

The use case is to alert users to permit modifications of behavior for the remainder of the day. The user values the first true positive prediction, but duplicate notifications for the same event produce no added benefit and are considered disruptive. The research team determined that the utility of positive predictions (precision) should be maximized at a recall of approximately 50% (predictions capture half the available benefit). Since this is a *nonconforming, alarm-centric scenario*, utility assignment functions were chosen identical to those used in scenario C (Fig. 5).

A ranging optimization was carried out to identify optimal settings for cutoff and snooze duration, with optimal results obtained from a selected cutoff of 25% and a snooze duration of 24 hours. For purposes of illustration, the process was repeated for a snooze duration of zero, with optimal results obtained at a selected cutoff of 27.5%.

Table X shows that snoozing improved precision and reduced unwanted alarms by 52% with no loss of recall; u-metrics match the user experience whereas c-metrics do not.

## VII. Discussion

We have shown that traditional count-based metrics cannot properly represent the real-world performance of Boolean classifiers in nonconforming workflows. Since a majority of clinical workflows are nonconforming, this is a severe limitation. A novel solution addresses the roots of the problem with a new system of utility-based metrics derived from a fundamentally different **utility-based confusion matrix** in which:

1. Each prediction is assigned a quantitative utility based on the state of the system and the workflow context for the prediction. Utility functions account for temporal, contextual, and other sources of variance in prediction utility.
2. The utility of a prediction may be adverse or beneficial regardless of whether the prediction itself was true or false.
3. The confusion matrix collates beneficial vs adverse utility rather than collating true vs false predictions.
4. The confusion matrix includes complementary utility elements representing missed benefit and avoided harm.

Others have recognized the shortcomings of traditional c-metrics for assessing the realized performance of Boolean classifiers, but they do not explain the precise conditions under which count-based metrics will fail nor the reasons why this occurs. Each prior effort mitigates some aspect of utility nonuniformity for a specific application in a specific domain, but none offers a universal model or general solution to the problem. Each applies corrections to statistical measures that still incorporate underlying assumptions of dichotomization and symmetric prediction utility, even when those assumptions do not hold. This can result in problems and confusion.

For example, when utility coefficients are applied to the predictions of a traditional count-based confusion matrix, the matrix cell for true positive predictions may receive contributions of both beneficial and adverse utility (e.g., when some true positive predictions have beneficial utility and others have adverse utility). This can cause derived metrics to lose their classic meaning. The metric *recall* (classically the fraction of total available benefit that is captured by positive predictions) is calculated as *(TP / (TP+FN)*. If utility coefficients cause the numerator to contain both beneficial and adverse utility, the formula no longer yields *recall*, but something entirely different. Similar problems result from utility corrections using prediction reclassification.

For example, Jung et al. applied a utility coefficient to each cell of the confusion matrix in a scenario in which the number of times a clinician could act on predictions of 12-month mortality was capped due to limited resources for advanced care planning. True positive predictions were retrospectively recategorized as false negative predictions whenever the original prediction could not be acted upon due to resource limitations [11]. Such a recategorization facilitates the selection of classifier cutoff (e.g., to maximize economic utility) but the resulting c-metrics are rendered confusing because they no longer reflect realized classifier performance: both recall and precision are falsely reduced by the reclassification. After the cap is reached, recall and precision will continue to be falsely reduced by additional "false negative" predictions, even though (1) no further value exists to be captured and (2) the actual predictions were neither false nor negative.

U-metrics avoid this and other difficulties because the u-confusion matrix is natively based on utility rather than prediction truth. U-metrics correctly reflect and explain the realized performance of Boolean predictors in nonconforming workflow, including situations in which some predictions cannot be acted upon because of resource limitations. They make it possible to characterize the real-world utility attributable to a predictor; improve and optimize predictor performance; retain the implications of familiar concepts such as precision, recall, and area under the precision-recall curve; and use quantitatively correct metrics to measure, assess, and compare classifier performance and the effects of classifiers on real-world workflows and outcomes.

Utility-based metrics may be enhanced and extended in the same manner as c-metrics. Additional attributes of a prediction may be calculated along with the primary utility score, providing a convenient way to assess the performance of a classifier from multiple perspectives simultaneously.

## VIII. Limitations

U-metrics accurately describe the performance of a Boolean predictor only to the extent that utility functions accurately estimate the actual utility delivered. In some scenarios it may be difficult to estimate the relative value of predictions made in different contexts. Nonetheless, utility-based metrics will outperform classic metrics for classifier performance in every situation *unless* the implied utility function inherent in c-metrics ($u = 1$ for all predictions) is a better fit for the workflow than the utility functions selected during implementation.

U-metrics correctly reflect the realized performance of Boolean predictors across nonconforming workflows where c-metrics do not. However, since utility metrics depend on context-specific utility values, it may still be difficult to compare models that operate in different contexts. Comparison may be facilitated through sharing of models and datasets, or by publishing metrics for a range of utility value strategies.

Although snoozing improves predictor performance in many cases, overlong snoozing could reduce performance and increase risk. We have described a *ranging* method to identify the best-performing combinations of cutoff, prediction frequency, and snooze duration.

## IX. Conclusion

A novel approach to Boolean classifier performance assessment uses utility-based *u-metrics* to account for contextual and temporal variation and other causes of nonuniform distribution of utility across predictions made in different contexts. Utility-based metrics correctly measure and reflect real-world performance in many nonconforming workflows where traditional count-based metrics fail, including the very common "alarm-centric" scenario in which notifications ("alarms") are triggered by positive predictions and multiple predictions are made for each event.

Snoozing algorithms may significantly improve realized predictor performance, particularly in alarm-centric scenarios. Utility-based metrics correctly measure the performance of predictors in such scenarios both with and without snoozing, whereas classic count-based metrics cannot. Practical methods exist to facilitate the design and implementation of snoozing algorithms in conjunction with utility-based metrics.

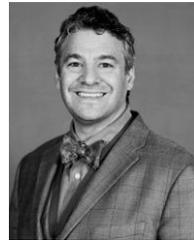

**Jonathan A. Handler, MD, FACEP, FAMIA** is a clinical informaticist and an emergency physician with board certifications in Clinical Informatics and Emergency Medicine. He is the Senior Fellow for Innovation at OSF Healthcare, the President of Keylog Solutions LLC, and an Adjunct Associate Professor at the Department of Emergency Medicine Northwestern University's Feinberg School of Medicine. He is a Fellow of the American Medical Informatics Association and a Fellow of the American College of Emergency Physicians.

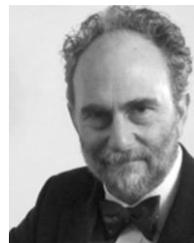

**Craig F. Feied, MD, FACEP, FAAEM, FAMIA, FACPh** is a biophysicist, clinical informaticist, and emergency physician with board certifications in Clinical Informatics and Emergency Medicine. He is the Chief Science Officer of Whispersom Corporation, Managing Director of Asatte LLC, and a Professor of Emergency Medicine at Georgetown University School of Medicine. He is a senior member of IEEE, a Fellow of the American Medical Informatics Association, a Fellow of the American College of Emergency Physicians, a Fellow of the American Academy of Emergency Physicians, and a Fellow of the American College of Phlebology.

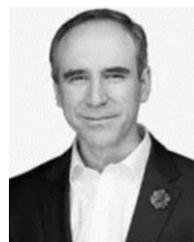

**Michael T. Gillam, MD, FACEP** is a clinical informaticist and an emergency physician with board certifications in Clinical Informatics and Emergency Medicine. He is a Senior Fellow at the MedStar Institute for Innovation, the Chief Digital Officer of Fountain Life, and the Chief Executive Officer of HealthLab. He is a Fellow of the American College of Emergency Physicians.